\def\year{2022}\relax
\documentclass[letterpaper]{article} 
\usepackage{aaai22}  
\usepackage{times}  
\usepackage{helvet}  
\usepackage{courier}  
\usepackage[hyphens]{url}  
\usepackage{graphicx} 
\usepackage{natbib}  
\usepackage{caption} 
\DeclareCaptionStyle{ruled}{labelfont=normalfont,labelsep=colon,strut=off} 
\usepackage{algorithm}
\usepackage{algorithmic}

\usepackage{newfloat}
\usepackage{listings}
\floatstyle{ruled}
\newfloat{listing}{tb}{lst}{}
\floatname{listing}{Listing}
\title{Diversity-based Deep Reinforcement Learning Towards Multidimensional Difficulty for Fighting Game AI}
\author{
    Emily Halina,
    Matthew Guzdial
}
\affiliations{
    Department of Computing Science, Alberta Machine Intelligence Institute (Amii)\\
    University of Alberta, Edmonton, Alberta, Canada\\
    \{ehalina, guzdial\}@ualberta.ca
}

\begin{document}

\maketitle

\begin{abstract}
In fighting games, individual players of the same skill level often exhibit distinct strategies from one another through their gameplay.
Despite this, the majority of AI agents for fighting games have only a single strategy for each ``level'' of difficulty.
To make AI opponents more human-like, we'd ideally like to see multiple different strategies at each level of difficulty, a concept we refer to as ``multidimensional'' difficulty.
In this paper, we introduce a diversity-based deep reinforcement learning approach for generating a set of agents of similar difficulty that utilize diverse strategies.
We find this approach outperforms a baseline trained with specialized, human-authored reward functions in both diversity and performance. 


\end{abstract}

\section{Introduction}




Fighting games have long featured AI agents that act as opponents for human players to play against.
The majority of these AI agents are created using a notion of ``linear'' difficulty, meaning the only distinction between agents is a difficulty rating in a fixed range from easy to hard.
Despite its prevalence, this model of linear difficulty is not well aligned with the average player's experience playing against other humans.
The high amount of player expression within fighting games allows for human players of similar skill levels to use the same mechanics in disparate ways to uniquely challenge their opponents \cite{dhami_2021}.
For example, one player may use a character's tools to keep their distance and chip away at an opponent, while another may use the same tools to aggressively approach the opponent in close quarters.
This disconnect can make players feel inadequately prepared for playing against other humans, and could cause some to drop a game entirely.

This problem could be mitigated through the use of a ``multidimensional'' difficulty system in which agents are distinguished by both linear difficulty and additional qualities such as playstyle.
To the best of our knowledge, this is the first academic work to identify this research problem.
By incorporating a notion of multidimensional difficulty, fighting games could provide players the experience of playing against multiple diverse strategies.
This could allow players to better prepare for playing against other human players and push them to more fully explore the mechanics of the game.



A major challenge in attaining multidimensional difficulty is the overall design burden of creating fighting game AI agents.
Fighting games agents often require significant hand-authoring to be effective, using rules-based systems or Finite State Machines (FSMs) to accommodate complex game mechanics and character interactions \cite{majchrzak2015advanced}.
Due to this difficulty, some games resort to ``cheating'' by reading the player's inputs to artificially increase the difficulty of an AI agent, further breaking parity between human and AI opponents.
There has been prior work in alleviating this designer burden through the use of Reinforcement Learning (RL) techniques to autonomously train AI agents to play fighting games \cite{kim2020mastering, oh2021creating}.
However, the majority of this work has been focused solely on playing the game, which is a testament to the difficulty of developing such agents.
As such, the problem of automatically developing agents that exhibit diverse strategies for fighting games has been relatively unexplored.




In this paper we focus on the task of training a group of agents of similar difficulty that utilize diverse strategies from one another.
Ideally, these agents would provide a more complete, robust gameplay experience for players while providing a suitable challenge.

Towards this goal, we propose Brisket, a diversity-based deep RL approach for learning a set of equally skilled, diverse strategies inspired by \cite{eysenbach2018diversity}'s Diversity is All You Need (DIAYN).
We implemented Brisket in FightingICE, a fighting game research platform built for the testing and evaluation of AI agents \cite{fightingice}.
We evaluated the policies learned by Brisket against a set of ``human-authored'' baseline agents trained with specialized, human-authored reward functions, and found that they outperformed these baseline agents in both effectiveness and diversity.
Therefore, we claim that a diversity-based RL approach can be an effective way to produce enemy AI agents for fighting games with multidimensional difficulty.




\section{Related Work}




\subsection{NPC Generation}
Methods of algorithmically generating NPC behaviours and mechanics have been developed for many game domains.
These methods fall between automated game playing and Procedural Content Generation (PCG), defined as the generation of game content through AI or algorithmic means \cite{hendrikx2013procedural}.
The task of NPC generation is often included in prior work through the generation of entire games or game mechanics \cite{conceptualexpansion, cook2016angelina}.
We instead focus on the generation of NPC behaviour with a predefined set of mechanics.

Prior systems to generate NPC behaviour generation have used constructive, search-based, and constraint-based PCG techniques \cite{barros2016playing, mitsis2020procedural}.
One such system introduced by \cite{bossgeneration} utilized program synthesis for the generation of unique boss behaviour with pre-defined constraints.
These existing methods do not focus on generating a diverse population of NPCs, but rather the generation of individual NPCs.
Quality-diversity search algorithms exist for the generation of such populations, and have been applied to level generation and 3D model creation \cite{gravina2019procedural}.
While to the best of our knowledge quality-diversity search has not been applied to NPC generation, this could represent an alternative approach to our problem.


\subsection{Fighting Game AI}

There has been a vast amount of prior work on the creation of AI agents for playing fighting games.
Some prior work focuses on utilizing hand-authored rules-based systems to create competitive agents with a high amount of designer controllability \cite{sato2015adaptive}. 
\cite{majchrzak2015advanced} introduced one such rules-based system that utilized dynamic scripting to adapt an AI to a given opponent by selecting rules from a human-authored corpus.
Rules-based systems can be specifically authored to achieve desired qualitative behaviour, such as in \cite{yuda2019creating}'s system for creating an ``affective'' agent that makes decisions based on an emotional appraisal engine.
One weakness of rules-based systems is the intensive amount of hand-authoring required by expert designers. 
In comparison, our method focuses on learning policies without direct human authoring of an agent's behaviour.

In recent years, Monte Carlo Tree Search (MCTS) and RL-based fighting game agents have become more dominant, with some agents defeating top rated human players \cite{kim2018hybrid, oh2021creating, kim2020mastering}.
The majority of these works focus on playing the game as well as possible as opposed to generating novel or diverse strategies.
A notable exception is \cite{ishii2018monte}'s system for incorporating human-authored NPC ``personas'' into an MCTS agent's decision making.
While this system created agents using the same architecture that utilized distinct strategies, these strategies required authoring in terms of their preferred actions.
Our human-authored baseline, described below, takes inspiration from this work.

\subsection{PCGRL}
Procudural Content Generation via Reinforcement Learning (PCGRL) refers to the generation of game content via the use of RL techniques \cite{khalifa2020pcgrl}.
Most prior PCGRL work focuses on level generation and puzzle design \cite{ zakaria2022procedural, delarosa2021mixed}.
Our approach, Brisket, takes inspiration from DIAYN, which in turn was inspired by prior diversity-based RL and evolutionary methods \cite{eysenbach2018diversity, gregor2016variational, such2017deep}.
The application of diversity-based reinforcement learning to the domain of NPC generation is novel to the best of our knowledge.
While Brisket is inspired by DIAYN, we modified the algorithm in order to better suit our domain and problem. 

\section{FightingICE}

\begin{figure}
    \centering
    \includegraphics[scale=0.33]{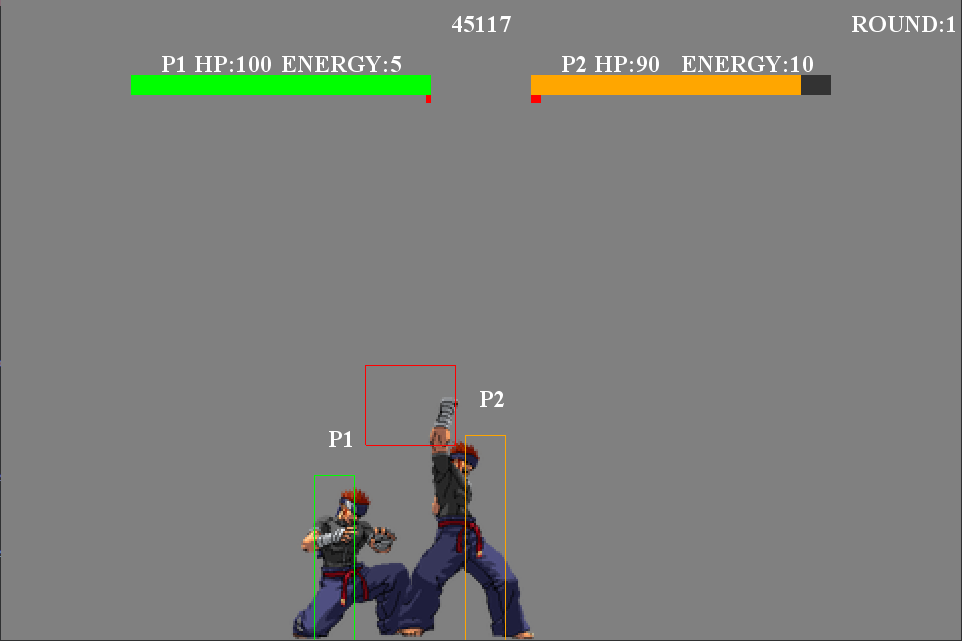}
    \caption{A screenshot of the FightingICE platform. Health and energy resources for each player are displayed at the top of the screen. 
    When an agent performs a move, a temporary red hitbox is created.
    This hitbox will damage the opponent if their hurtbox intersects with it.
    }
    \label{fig:fightingice}
\end{figure}

In this section we briefly introduce FightingICE, the environment we used as a testbed for our approach Brisket.
FightingICE is an open-source research platform built for the development and evaluation of AI agents for fighting games \cite{fightingice}.
We chose to use FightingICE due to the platform's support of AI development, its relative simplicity in comparison to other fighting games, and its proven success as a research platform \cite{chen2021interpretable, ishihara2016applying}

Figure \ref{fig:fightingice} depicts a screenshot of a match being played between two AI agents in FightingICE.
FightingICE is a simple 2D fighting game, where characters have access to a variety of punches, kicks, and basic evasive maneuvers.
Characters also have access to special moves which consume energy, a limited resource gained from hitting the opponent or being hit.
FightingICE currently contains three characters, but for simplicity we restrict our focus to the default character ZEN, as in prior work \cite{chen2021interpretable}.
Brisket was developed and tested on version 4.50 of the platform.

\begin{figure*}[h]
    \centering
    \includegraphics[scale=0.45]{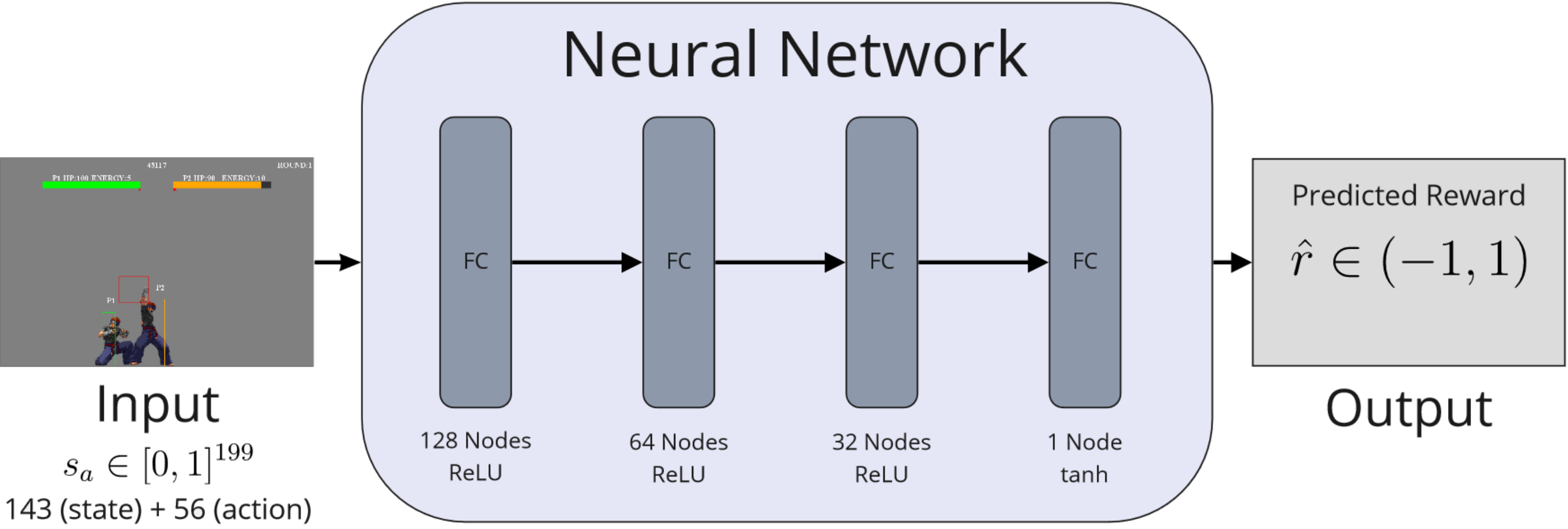}
    \caption{A visualization of the DQN agent architecture. The current state $s$ and proposed action $a$ are concatenated as input and passed through a fully connected neural network, which predicts the expected reward $\hat r$ of taking action $a$ in state $s$.}
    \label{fig:arch_figure}
\end{figure*}

Notably, because FightingICE was designed for the comparison of automated game playing approaches, it was not a trivial task to use it for NPC behaviour research.
Multiple adjustments had to be made to the platform in order for it to function for Brisket, the most notable of which is the removal of a 15 frame input delay.
We provide the adjusted source code in a GitHub repository, along with the implementation of Brisket in FightingICE\footnote{\url{https://github.com/emily-halina/brisket}}.

\section{System Overview}



%
In this section we discuss the technical details of Brisket, our diversity-based deep RL approach.
On a high level, the approach involves training multiple agents concurrently, rewarding each agent based on their diversity from one another according to a discriminator.
We then fine-tune these agents individually on a general reward function with the goal of creating agents of a similar difficulty that are diverse from one another.

We begin by discussing the formulation of the Markov Decision Process (MDP) for our task, describing the state and action spaces and reward function.
We then detail the architecture used by our deep Q-agents and discriminator.
The Diversity-Based Learning subsection covers the diversity step of our training approach, outlining the algorithm used for training our agents ``without a reward function,'' and the Fine-tuning subsection covers the fine-tuning step of our training approach.

\subsection{Markov Decision Process}

An MDP is defined as a state space $S$, action space $A$, and reward function $R$. We ignore the transition function as we have deterministic actions. The goal of an RL approach is to learn a policy function $\pi_{R} : S \rightarrow A$ which maps a given state to the optimal action to take in that state to maximize value according to $R$.

There is no default state representation in FightingICE, and so we had to design our own.
Our state representation contains 143 variables, all of which are floats normalized between 0 and 1.
The first 14 variables contain basic information about both players, including current health, current energy for special moves, position, and velocity.
The following 112 variables represent the current status of each player, which includes options such as standing, crouching, or performing a move.
This status is encoded as a one-hot vector, with each status being associated with a unique action.
We include the remaining time in the round as a single variable to allow for more nuanced decision making, such as an agent becoming more aggressive or defensive at the end of the round.
The final 12 variables encode the relative position and hit damage of the two most recent projectiles from both players.
Projectiles persist between states, continuing to move until colliding with a player or reaching the end of their range.
We chose to include only the most recent two projectiles in an effort to reduce the size of each state, as it is very rare for more to be active at any given time.
When there are no projectiles or only one projectile, the corresponding unused variables are given a placeholder value of 0.



FightingICE's default action representation was a good fit for this task, and so we can just enumerate all possible actions for the action space of the MDP.
There are 56 possible actions agents can take, including simple options such as walking forward or punching, along with complex inputs such as projectiles or special moves.
These actions were encoded as an one-hot vector of length 56.


Brisket uses two separate reward functions: the fine-tuning function $R_f$ and the diversity-based function $R_d$.
$R_d$ is defined and discussed below.
$R_f$ is defined as
\begin{equation}
    R_f(s, a) = T(s) \cdot 1000 
\end{equation}
where $T(s) = 1$ if $s$ is a positive terminal state (winning the game), $T(s) = -1$ if $s$ is a negative terminal state (losing the game), and $T(s) = 0$ otherwise.
We chose not to include an existential penalty as not to invalidate strategies that may attempt to win by time-out, as the state representation includes the remaining time in a round.
Additional specialized reward functions used for the development of our baseline DQN agents are defined in the Evaluation section.

\subsection{Architecture}



Our approach Brisket involves the training of a fixed number of Deep Q-Network (DQN) agents in tandem.
DQN agents utilize a neural network to approximate a lookup table mapping a given state-action pair to the expected value of taking a given action in that state \cite{li2017deep}.
In this subsection, we focus on the architecture used by each DQN agent, as well as the architecture used for the discriminator which determines the reward in the diversity-based training step.

Figure \ref{fig:arch_figure} depicts a visualization of the DQN architecture.
The model takes in a length 143 vector representing a state $s$, along with a length 56 one-hot vector representing the proposed action $a$ to evaluate.
These vectors are concatenated and passed through 3 fully connected layers of 128, 64, and 32 neurons respectively.
Each of these layers use rectified linear unit (ReLU) as the activation function.
Our final output layer is a single node using the tanh activation function to represent a predicted reward $\hat r \in (-1, 1)$.
The use of ReLU and the parameters of our model are informed by prior work on DQN agents for the fightingICE domain, with the notable change of predicting a reward based on a state-action pair rather than treating the model as a 56-class classification problem \cite{takano2018applying}.
This change was to reduce the complexity of the model's representation, allowing the model to more easily learn relationships between states and individual actions.

Brisket uses a discriminator during the diversity step in order to reward agents based on their diversity.
This discriminator takes in a state-action pair as input and predicts the likelihood that this pair came from each of our agents.
This likelihood is then used in the reward function for these agents, encouraging them to take more and more diverse actions from one another.
The discriminator uses the same architecture as each DQN agent, with the only architectural change being at the output layer.
The output layer has $K$ nodes, where $K$ is the number of policies to learn, and uses the softmax activation function in place of tanh.
We chose to use the same architecture for both the discriminator and DQN agents, as they are modelling similar relationships between state-action pairs and their desired output.

\subsection{Diversity-based Learning}

\begin{algorithm}[t]
\small
\caption{Diversity-based learning algorithm}
\label{algo:div}
\begin{algorithmic}[l]
\STATE Let $S_\pi$ be fixed set of policies, $D$ discriminator
\FOR{episode $e$}
\FOR{round $r$}
\STATE Sample $\pi_r \sim S_\pi$ to make choices during $r$
\FOR{timestep $t$}
\STATE $a_t \leftarrow \epsilon-greedy(\pi_r, s_t)$
\FOR{policy $\pi \in S_\pi$}
\STATE // calculate reward based on $D$
\STATE $r_{\pi, t} \leftarrow R_d(s_t, a_t, \pi, D)$  
\ENDFOR
\STATE Step environment $s_{t+1} \leftarrow (s_t, a_t)$
\ENDFOR
\ENDFOR
\STATE Update each policy $\pi \in S_\pi$ with samples from $e$
\STATE Update discriminator $D$ with samples from $e$
\ENDFOR
\end{algorithmic}
\end{algorithm}


Brisket is largely inspired by DIAYN, an approach for concurrently training a group of RL agents which are rewarded based on a discriminator which determines how distinct these agents are from one another
\cite{eysenbach2018diversity}.
Brisket makes the major change of using state-action input for our DQN agents and discriminator rather than just state \cite{eysenbach2018diversity}.
This change was due to our task of behaviour generation, which differs from the exploration task for which DIAYN was originally designed.

Algorithm \ref{algo:div} describes the procedure for learning a set of policies that are diverse from one another according to a discriminator.
We begin by defining $S_\pi$ as the fixed set of policies to be learned, and $D$ as the discriminator.
Each episode contains a fixed number of ``rounds'' in fightingICE, which can be thought of as individual rollouts from an initial state.
For each of these rounds, we sample a policy $\pi_r \sim S_\pi$ to ``pilot'' the given round.
This is done in an attempt to balance the training data for the discriminator, as well as to allow for the possibility of encountering states which are only feasibly reachable through following the same policy for an entire round.
For each timestep, we select an action according to $\pi_r$ using $\epsilon$-greedy, with $\epsilon$ being annealed from $0.95$ to $0.05$ linearly across 50 episodes to ensure a high amount of early exploration.
We then calculate the reward for each policy $\pi$ using the diversity-reward function $R_d$, defined as
\begin{equation}
    R_d(s_t, a_t, \pi, D) = log D(s_t, a_t | \pi) - log \frac{1}{|S_\pi|}
\end{equation}
where $log$ is the natural logarithm, and $D(s_t, a_t | \pi)$ is the discriminator $D$'s prediction that the state-action pair $(s_t, a_t)$ came from policy $\pi$.
We then step the environment forward using $a_t$, collecting training samples for the DQN agents and discriminator.
Each of the rounds are played against a ``random agent'' which selects actions uniformly at random. 
This was to allow for the policies to see as many unique states as possible.
At the end of each episode, we update each policy based on their respective samples, then update the discriminator based on the true pilot for each state-action pair.
We found experimentally that we did not need to use a fixed replay buffer in this environment, as supported by prior work which showed large replay buffers are not universally beneficial \cite{fedus2020revisiting}.
We perform an 80-10-10 train-val-test split when training the discriminator, recording the validation accuracy to check for convergence.
The DQN agents and discriminator are both trained using the Adam optimizer, with MSE and categorical cross-entropy as loss functions, respectively.
For each of these updates, we train for 5 epochs with a learning rate of $10^{-5}$ and a batch size of 1.
Our selection of hyperparameters was informed by prior work \cite{halina2021taikonation}.

While Brisket allows for an arbitrary number of learned policies, for our evaluation we trained three agents concurrently using this strategy.
This small number was chosen to evaluate if diverse, effective strategies could be learned without an additional step of ``pruning'' a number of ineffective strategies.
After 50 episodes with 100 rounds of gameplay in each episode, we converged, reaching a discriminator test accuracy of $99.5\%$.
This took approximately 48 hours to run on a single machine using an AMD Ryzen 5 3600x CPU and NVIDIA GeForce 1080 Ti GPU.
This suggests Brisket may be accessible to those without substantial computing power, such as fighting game developers.

\subsection{Fine-tuning}

In the second step of our approach, each of the learned policies are individually fine-tuned to a similar level of difficulty.
This is to ensure each of the policies is not trivially easy, as we did not reward winning in the previous step.
For each of these policies, we trained for an additional 50 episodes, using the fine-tuning reward function $R_f$ defined previously.
During this step, we lower the learning rate to $10^{-6}$ and use a fixed $\epsilon$ of 0.05 in an attempt to not drastically alter the pre-existing learned behaviour of the policies.
As in our diversity step, these rounds are played against a random agent to learn the most general strategies possible.

Between every episode we had the agent play 9 rounds (3 matches of 3 rounds) against the random agent using greedy action selection, recording the average reward for these rounds.
After 50 episodes with 100 rounds of gameplay in each, our reward plateaued and we stopped training.
Finally, we selected the highest scoring episode in terms of average reward as the final version of the policy.
This was to ensure we selected the most effective version of each policy.

By running Brisket once, we created three diverse, effective agents that utilize differing strategies to win games.
We named these agents ``Combo,'' ``Rushdown,'' and ``Sweeper'' based on their behaviour, which is overviewed in the Agent Descriptions section.
The top performing version of each agent came from fine-tuning episodes 17, 6, and 20 respectively.
We chose to only run Brisket once as an initial test, and to avoid the possibility of stumbling into a good set of agents by the law of large numbers.

\section{Evaluation}

In this section we discuss the evaluation criteria and the human-authored baseline we compare Brisket against. 
Our goal is to create a set of multidimensionally difficult agents of similar skill levels that exhibit distinct strategies from one another.
As such, we require another set of agents to compare against, as well as methods for measuring both capability and diversity of strategy.

As a baseline we used three DQN agents employing hand-authored reward functions that were designed and trained prior to our diversity-based approach.
This was to avoid any chance of the output agents from Brisket biasing the design of the baseline reward functions.
These reward functions were designed with the intention of eliciting desired, diverse behaviours while still optimizing for optimal play.
We chose to compare against DQN agents instead of other human-authored approaches such as rules-based systems to make the comparison as balanced as possible.
Therefore, this comparison can be seen as comparing Brisket against our own ability to create diverse RL agents. 
We named these three agents ``Aggressive,'' ``Balanced,'' and ``Counter'' after the behaviours we intended for them to exhibit.
The top performing version of each agent came from episodes 76, 64, and 99 respectively.
The training of these agents and their reward functions are outlined in the Baseline DQN Training subsection, and their behaviours are summarized in the Agent Descriptions section. 

We performed two different evaluations comparing the policies learned by Brisket to the baseline agents.
The first was a measurement of diversity across randomized states.
To accomplish this, we collected ten thousand random states by pitting two random agents against each other.
We then queried each of our six agents to greedily select the action they would take in each of these states, and recorded their choices.
We then looked at the percentage of states in which each agent chose a distinct action compared to each other agent.
This gives us a notion of how diverse the actions of the agents were relative to one another across a wide variety of states.

The second evaluation was a head-to-head round-robin tournament in which each agent played 10 matches against each other agent.
We chose to run for only 10 matches as we found running additional games produced consistent results.
In half of these matches, each agent started on the left side, and in the other half the right to avoid giving any agent an unfair advantage.
We recorded the win-loss-tie records of each agent in every match-up, as well as total records.
This evaluation is intended to measure the effectiveness of each set of agents in relation to one another, as well as ensuring we are not sacrificing effectiveness for diversity with Brisket.
If Brisket produces more diverse strategies of a consistent difficulty, this would represent strong evidence towards it being able to achieve multidimensional difficulty.

\begin{table*}[h]
\centering
\small
\begin{tabular}{lccc|ccc|c}
\hline
                       & \textbf{Combo} & \textbf{Rushdown} & \textbf{Sweeper} & Aggressive & Balanced & Counter & All \\ \hline
\textbf{Combo}       & 0\%              & 98.59\%             & 99.12\%            & 99.53\%                & 96.23\%              & 99.75\%             & 93.59\%      \\ \hline
\textbf{Rushdown}    & 98.59\%          & 0\%                 & 83.87\%            & 99.04\%                & 97.75\%              & 98.74\%             & 78.73\%      \\ \hline
\textbf{Sweeper}     & 99.12\%          & 83.87\%             & 0\%                & 99.94\%                & 99.74\%              & 99.96\%             & 82.69\%      \\ \hline
\textbf{Aggressive} & 99.53\%          & 99.04\%             & 99.94\%            & 0\%                    & 87.89\%              & 94.73\%             & 83.19\%      \\ \hline
\textbf{Balance}    & 96.23\%          & 97.75\%             & 99.74\%            & 87.89\%                & 0\%                  & 78.59\%             & 62.40\%      \\ \hline
\textbf{Counter}    & 99.75\%          & 98.74\%             & 99.96\%            & 94.73\%                & 78.59\%              & 0\%                 & 73.75\%      \\ \hline
\end{tabular}
\caption{The results of the diversity comparison across ten thousand random states. 
The ``All'' columns depicts the percentage of actions taken in a state that were diverse from all other agents. Each cell represents the diversity between two specific agents. 
Combo, Rushdown, and Sweeper are the agents learned by Brisket.}
\label{diversitytable}
\end{table*}

\begin{table*}[h]
\centering
\small
\begin{tabular}{lccc|ccc|c}
\hline
\multicolumn{1}{c}{} & \textbf{Combo} & \textbf{Rushdown} & \textbf{Sweeper} & Aggressive & Balanced & Counter & Overall Record \\ \hline
\textbf{Combo}       & X                & 10 - 0 - 0          & 10 - 0 - 0         & 9 - 1 - 0           & 5 - 5 - 0         & 8 - 2 - 0        & 42 - 8 - 0              \\ \hline
\textbf{Rushdown}    & 0 - 10 - 0       & X                   & 4 - 1 - 5          & 5 - 5 - 0           & 4 - 6 - 0         & 5 - 5 - 0        & 18 - 27 - 5             \\ \hline
\textbf{Sweeper}     & 0 - 10 - 0       & 1 - 4 - 5           & X                  & 9 - 1 - 0           & 5 - 5 - 0         & 10 - 0 - 0       & 25 - 20 - 5             \\ \hline
\textbf{Aggressive} & 1 - 9 - 0        & 5 - 5 - 0           & 1 - 9 - 0          & X                   & 5 - 5 - 0         & 0 - 10 - 0       & 12 - 38 - 0             \\ \hline
\textbf{Balance}    & 5 - 5 - 0        & 6 - 4 - 0           & 5 - 5 - 0          & 5 - 5 - 0           & X                 & 5 - 5 - 0        & 26 - 24 - 0             \\ \hline
\textbf{Counter}    & 2 - 8 - 0        & 5 - 5 - 0           & 0 - 10 - 0         & 10 - 0 - 0          & 5 - 5 - 0         & X                & 22 - 28 - 0             \\ \hline
\end{tabular}
\caption{The results of the head-to-head round-robin tournament, listed by win - loss - tie from the perspective of the agent in each row. Combo, Rushdown, and Sweeper are the agents learned by Brisket.}
\label{tournamenttable}
\end{table*}

\subsection{Baseline DQN Training}
In this subsection we discuss the training procedure for our baseline DQN agents, along with their individual reward functions.
We kept as many parameters consistent between Brisket and the baseline as possible while training.
Each agent used the same architecture described previously.
We used the same training time, hyperparameters, and final policy selection as in Brisket.
This consistency was intended to reduce the factors affecting the agent's performance beyond the specific change of the reward function.

The reward function for each agent includes the use of $T(s)$ as defined in Section 4.1 to reward agents for winning or losing games, along with an existential penalty of $p = 1$ to keep agents from stalling or getting stuck in ``loops'' of actions.
We found experimentally that this penalty was necessary to achieve reasonable behaviour from each agent.
These reward functions are as follows.

\subsubsection{Aggressive Agent:} The ``Aggressive'' agent uses the reward function $R_a$ designed to bias the agent's behaviour towards hitting the opponent above all else.
$R_a$ is defined as
\begin{equation}
    R_a(s, a) = T(s) \cdot 1000 + A(s) \cdot 100 - p
\end{equation}
where $A(s) = 1$ if the agent dealt damage to the opponent in $s$, and $A(s) = 0$ otherwise.

\subsubsection{Balanced Agent:} The ``Balanced'' agent uses the reward function $R_b$ designed to make the agent take into account both offense and defense when making decisions.
$R_b$ is defined as
\begin{equation}
    R_b(s, a) = T(s) \cdot 1000 + A(s) \cdot 100 - B(s) \cdot 50 - p
\end{equation}
where $A(s)$ is as above, and $B(s) = 1$ if the agent was dealt damage in $s$, and $B(s) = 0$ otherwise.
$B(s)$ is weighted lower than $A(s)$, as we found experimentally that the agent would get trapped in local maxima with equal weighting.

\subsubsection{Counter Agent:} The ``Counter'' agent uses the reward function $R_c$ designed to bias the agent into ``counter-attacking'' the opponent while they are in the middle of a move.
$R_c$ is defined as
\begin{equation}
    R_c(s, a) = T(s) \cdot 1000 + C(s) \cdot 100 - p
\end{equation}
where $C(s) = 1$ if the agent dealt damage to an opponent that was mid-move in $s$, and $C(s) = 0$ otherwise.

\section{Agent Descriptions}

Because we cannot realistically give all the policies for each agent, in this section we briefly describe each agent's behaviour from our perspective. We also provide a video of
the gameplay of each agent for additional context\footnote{\url{https://www.youtube.com/watch?v=GnoURvbHMLY}}.

Beginning with the learned diversity skills, \textbf{Combo} is an agent characterized by its chaining of fast moves together.
The agent uses a wide variety of moves depending on the situation, and has learned a difficult-to-escape combo that traps the opponent in the corner.
\textbf{Rushdown} is an agent which quickly approaches the opponent to relentlessly attack.
The agent frequently switches between high and low attacks, making it difficult for an opponent to effectively block or defend themself.
\textbf{Sweeper} is an agent that chooses to use a low sweeping kick in almost every circumstance.
The agent has learned to effectively chain these sweeps together, keeping the enemy permanently knocked down unless they correctly block or evade the attack.

The three human-authored skills all converged to similar maxima, with small distinctions between each policy.
This maxima involves using a wide-reaching uppercut move during the beginning of a round, followed by throwing projectiles until the agent is out of energy or the round is finished.
The \textbf{Aggressive} agent is the simplest form of this ``uppercut'' strategy, attempting to chain uppercuts against the opponent without caring for self-preservation.
The \textbf{Balanced} agent occasionally mixes up this base strategy with faster kicks, and has a higher preference toward throwing projectiles than the other human-authored agents.
We hypothesize that this is due to the notion of self-preservation in this agent's reward function.
The \textbf{Counter} agent also tweaks this base strategy by focusing more on uppercutting than throwing projectiles.
Instead of using energy on throwing projectiles, this agent performs a sliding kick move that goes under most other moves, allowing the agent to effectively ``counter hit'' the opponent mid-move as desired by the design of its reward function.

\section{Results}

In this section we discuss the results of the action diversity and head-to-head evaluations.



Table \ref{diversitytable} compares the diversity between each agent's chosen actions across ten thousand random states.
Overall, the agents learned by Brisket took actions that were diverse from the entire population of agents more often than the human-authored baseline agents.
This suggests the diversity-based agents are exhibiting more distinct behaviours than the baseline agents.
In particular, the Combo agent exhibited the most diverse behaviour, selecting actions that were diverse from the entire population 93.59\% of the time.
We hypothesize this is due to the high variety of moves used by the agent, as described in Section 6.
By comparison, the human-authored agents took less diverse actions, which we hypothesize is a result of the agents converging to similar local maxima despite the differences in their reward functions.
This may suggest a weakness in simultaneously optimizing towards desired diverse behaviours and optimal play, which Brisket subverts through its two-step process.
Notably, while the Aggressive agent outperforms the Rushdown and Sweeper agents in total diversity, this is primarily due to the differences between it and the diversity-based agents.
From our perspective, Brisket's agents are noticeably more distinct from one another than the human-authored agents, which highlights the difficulty of measuring human-subjective diversity without human evaluation.


Table \ref{tournamenttable} depicts the results of the head-to-head tournament between the six DQN agents.
Overall, Brisket outperformed the human-authored baseline in the tournament, with cumulative records of 85 wins - 55 losses - 10 ties and 60 wins - 90 losses - 0 ties respectively across both groups of agents.
This performance is promising, as it suggests that we did not sacrifice effectiveness for the sake of diverse behaviour.
The groups of agents showed a similar degree of internal balance, with each having two equally scoring agents, and one outlier.
Notably, Combo and Sweeper performed very similarly versus the human agents despite being almost completely distinct from one another in the diversity experiment.
This is a promising sign toward our goal of attaining multidimensional difficulty with Brisket.

\section{Discussion}
In this section we reflect upon our results, addressing the limitations of our evaluation and system design and discussing potential avenues for future work.

\subsection{Limitations}
The most notable limitation of our work is the lack of human evaluation.
While the agents generated by Brisket outperformed the baseline on our evaluation metrics, the lack of human evaluation makes it difficult to draw strong conclusions about the quality or effectiveness of these agents.
In particular, the results of our diversity evaluation highlight the difficulty in measuring something as qualitative and subjective as diversity.
As well, human evaluation would be helpful for evaluating the robustness of the learned agents, as there is a possibility the current strategies may be brittle or ineffective against human opponents.
While human evaluation is outside the scope of this paper, we feel it is a clear next step to guide future development of Brisket.

As well, there are a number of limitations concerning our evaluation \& result.
Both of the evaluations have relatively small sample sizes, which could present a potential source of bias.
However, in both the case of the diversity evaluation and head-to-head tournament, we found that using additional samples yielded almost identical results.
As well, in this paper we only utilized Brisket to learn a small number of agents due to computational constraints.
While Brisket may potentially be more effective with a larger number of learned policies, we interpret our results as a positive sign that the system may be useful to developers without substantial computing power.




\subsection{Future Work}
While in this paper we present one potential method for creating agents that exhibit multidimensional difficulty, there are other possible avenues for future exploration.
One such avenue could be the use of a state-based diversity metric in training which rewards agents based on the diversity of game states each agents encounters.
This may yield more diverse behaviour than our current action-based diversity metric, and the two could potentially be used in tandem to create agents which drive a match towards distinct states using disparate actions from one another.
As well, quality-diversity methods could represent another solution to this problem, and have yet to be applied in the domain of behaviour generation to our knowledge \cite{gravina2019procedural}.

\section{Conclusions}
In this paper we presented Brisket, a diversity-based deep RL approach with the goal of generating multidimensionally difficult fighting game agents.
We implemented Brisket in FightingICE, generating three agents to compare against a set of DQN agents using human-authored reward functions.
We found Brisket outperformed this human-authored baseline in both diversity and effectiveness.
We hope Brisket can represent a preliminary step towards multidimensional difficulty in games, and inspire other work to explore the problem further.

\section{Acknowledgements}
We acknowledge the support of the Alberta Machine Intelligence Institute (Amii).

\bibliography{aaai22.bib}

\end{document}